\documentclass[letterpaper, 10 pt, conference]{ieeeconf}

\usepackage{amsmath}
\usepackage{amssymb}
\usepackage{mathtools}

\usepackage{tabularx}
\usepackage{graphicx}
\usepackage{bm}
\usepackage{color}
\usepackage{float}
\usepackage{units}
\usepackage{url}
\usepackage{hyperref}
\usepackage{balance}
\usepackage{amsmath}
\usepackage{makecell}
\usepackage{cite}

\makeatletter 
\def\endfigure{\end@float} 
\def\endtable{\end@float}
\makeatother

\usepackage{subcaption}
\usepackage[]{graphicx}
\usepackage{caption}

\renewcommand{\unit}[1]{{\rm #1} }

\IEEEoverridecommandlockouts

\begin{document} 

\title{\Large \bf 			
Adapting Gait Frequency for Posture-regulating Humanoid Push-recovery via Hierarchical Model Predictive Control 
}

\author{Junheng Li$^*$, Zhanhao Le$^*$, Junchao Ma, and Quan Nguyen\thanks{ $^*$Equal contribution.}\thanks{All authors are with the Department of Aerospace and Mechanical Engineering, University of Southern California, Los Angeles, CA 90089.}\thanks{Email:{\tt\small $
\{$junhengl,zle,junchaom,quann$\}$@usc.edu}}\thanks{This work is supported by USC Departmental Startup Fund.}}

\maketitle


\begin{abstract}
Current humanoid push-recovery strategies often use whole-body motion, yet they tend to overlook posture regulation. For instance, in manipulation tasks, the upper body may need to stay upright and have minimal recovery displacement. This paper introduces a novel approach to enhancing humanoid push-recovery performance under unknown disturbances and regulating body posture by tailoring the recovery stepping strategy. We propose a hierarchical-MPC-based scheme that analyzes and detects instability in the prediction window and quickly recovers through adapting gait frequency. Our approach integrates a high-level nonlinear MPC, a posture-aware gait frequency adaptation planner, and a low-level convex locomotion MPC. The planners predict the center of mass (CoM) state trajectories that can be assessed for precursors of potential instability and posture deviation. 
In simulation, we demonstrate improved maximum recoverable impulse by 131$\%$ on average compared with baseline approaches.
In hardware experiments, a 125 ms advancement in recovery stepping timing/reflex has been observed with the proposed approach. We also demonstrate improved push-recovery performance and minimized body attitude change under 0.2 rad. 

\end{abstract}


\section{Introduction}
\label{sec:Introduction}

Humanoid robots possess the inherent advantage of generating human-like movements, enabling them to perform complex real-world tasks. 
However, achieving this goal presents a significant challenge, as humanoid robots must overcome their inherent instability caused by under-actuation and small support regions, which is further intensified in \text{manipulation} tasks with objects. A key area to explore is developing re-balancing control strategies to counteract \textit{unknown} external disturbances while maintaining the integrity of upper body tasks, such as in dynamic \textit{loco-manipulation} \cite{purushottam2023dynamic, foster2024physically, li2023hector}, where body posture regulation is a crucial aspect.

Traditionally, the push-recovery strategies on humanoid robots have been categorized as \textit{ankle}, \textit{hip}, or \textit{stepping} strategies \cite{aftab2012ankle,kanamiya2010ankle,kim2023model},  closely mimicking the natural human movements under external disturbances. The \textit{ankle} strategy typically leverages Zero Moment Point (ZMP) for stability \cite{nakaura2002balance,nenchev2008ankle}. The \textit{hip} strategy leverages the movement of the upper body with respect to the hip joint to counter the sudden change of angular momentum \cite{nenchev2008ankle,stephens2010dynamic,li2015push}. The \textit{stepping} strategies usually involve determining the Capture Point (CP) of the next step \cite{pratt2006capture, kim2023foot, shafiee2016push}.  

Recently, push-recovery strategies that involve arm motions enable improved stability through upper-body and hip motions \cite{miyata2019walking, khazoom2022humanoid}. However, excessive reliance on the upper body may cause large attitude changes and compromise the potential of upper body manipulation tasks.
Alternatively, reinforcement learning based approaches have demonstrated remarkable push-recovery control capabilities \cite{van2024revisiting,ferigo2021emergence,duburcq2022reactive}. A minimally constraining reward function is designed in \cite{van2024revisiting}, which enables emergent push-recovery behavior that involves taking a number of small steps to regain balance. Moreover, maintaining balance through adapting footstep placement strategies has seen in \cite{ding2022orientation, bang2024rl, chen2024learning}.
In this paper, we develop a model-based push-recovery \textit{stepping} strategy for humanoid robots that aims to strategically adapt \textit{gait frequency} to (i) improve push-recovery performance over some popular model-based baseline strategies and (ii) regulate body posture during recovery.

\begin{figure}[!t]
\vspace{0.2cm}
    \center	
    \includegraphics[clip, trim=0cm 10cm 0.3cm 0cm, width=1\columnwidth]{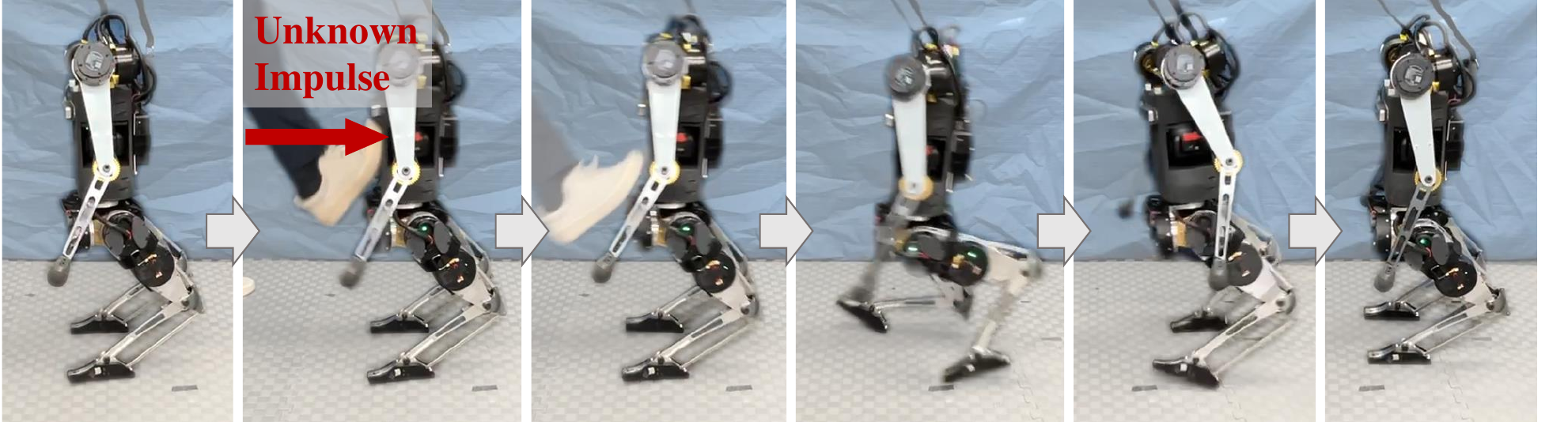}
    \caption{{\bfseries Push-recovery Experiment Snapshots on HECTOR V2.} Robot rebalances after a kick. Full experiment video: \url{https://youtu.be/pUdVy0RSaiE}}
    \label{fig:title}
    \vspace{-0.5cm}
\end{figure}

\begin{figure*}[!t]
\vspace{0.2cm}
		\center
		\includegraphics[clip, trim=0.75cm 0.3cm 0.85cm 0cm, width=2\columnwidth]{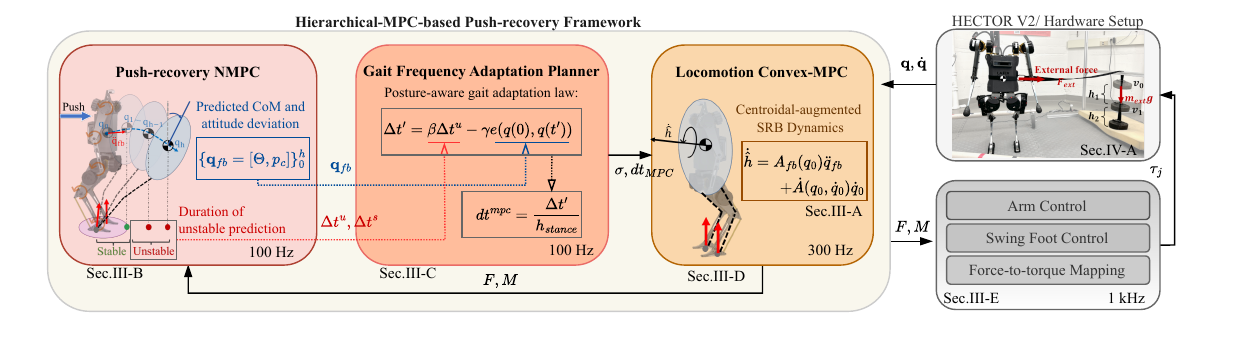}
		\caption{{\bfseries System Architecture} In high-hierarchy push-recovery NMPC, we employ a hierarchical dynamics model, fusing whole-body dynamics (WBD) in the first horizon and reduced-order dynamics in the rest. The WBD captures the effect of external impulse on float-base acceleration $\ddot{\mathbf q}_{fb}$, providing an accurate state evolution to start with. Our gait frequnecy adaptation planner gathers the unstable duration $\Delta t^u$ in NMPC prediction and associated body attitude deviations, determining a tailored stepping frequency in terms of MPC step duration and adapting contact sequence $\sigma$ for locomotion MPC to follow. }
		\label{fig:controlArchi}
		\vspace{-0.2cm}
\end{figure*}

Model-predictive-control (MPC) based approaches are widely favored in humanoid push-recovery, particularly those that involve \textit{stepping} strategies. A linear MPC in \cite{kim2023foot} leverages linear inverted pendulum model (LIPM) and CP-ZMP dynamics to scale the double support time between swings to enable more robust push recovery. A LIPM-based CP planning is modulated with ZMP and Centroidal Moment Pivot as control inputs in MPC \cite{shafiee2016push}. Moreover, a nonlinear divergent component of motion error is considered in a nonlinear MPC (NMPC) \cite{choe2023seamless}. The controller optimizes for both next-step location and duration, enabling seamless transitions of push-recovery strategies.  In \cite{gu2024robust}, a signal-temporal-logic-guided MPC is proposed to systematically reason the locomotion task and push-recovery task. The MIT humanoid showcased impressive stepping push-recovery with whole-body MPC with fixed stepping frequencies \cite{khazoom2024tailoring}. 

In our work, we leverage a hierarchical MPC-based framework with force-based control and dynamic stepping strategies without intermediate double-support phases. Our high-level NMPC utilizes a mixed-fidelity modeling strategy to predict the CoM state trajectory while adapting stepping timing and duration through heuristic-based gait frequency adaptation laws.
Hence, the hierarchical feature is twofold. First, we integrate push-recovery NMPC and locomotion MPC in a hierarchical fashion. The lower-level locomotion MPC utilizes tailored stepping frequency provided by the upper-level planners. In return, the upper-level NMPC adapts by incorporating the control inputs from the locomotion MPC for prediction. The second hierarchical feature exists in the modeling choice in push-recovery NMPC, which leverages a hierarchical modeling strategy \cite{li2021model, liu2022design}, blending whole-body dynamics (WBD) with newly proposed centroidal-augmented single rigid-body (CSRB) dynamics and full kinematics constraints. The WBD at the first time step captures the effect of unknown external disturbance on robot states for a more informed prediction.

The main contributions of this work are as follows:
\begin{itemize}
    \item We propose a hierarchical-MPC-based framework for humanoid push-recovery with \textit{stepping} strategy for (i) improved performance and (ii) body posture regulation. 
    \item The proposed hierarchical NMPC leverages whole-body dynamics and a newly proposed CSRB dynamics to predict potential instability and body attitude deviations under unknown disturbances.
    \item We design a posture-aware gait frequency adaptation law based on NMPC's CoM prediction and body attitude deviation to achieve posture-aware push-recovery by adjusting the stepping frequency.

    \item In both numerical and hardware experiments, we provide a comparative analysis to demonstrate the significance of gait frequency adaptation in push-recovery.    

\end{itemize}

The rest of the paper is organized as follows. Section  \ref{sec:overview} presents the overview of the proposed control system architecture. Section \ref{sec:approach} introduces the details of the humanoid robot dynamics modeling, the proposed push-recovery NMPC, the posture-aware gait frequency adaptation planner, and the locomotion MPC. Section \ref{sec:Results} presents the numerical comparative analysis and experimental validations.

\section{System Overview}
\label{sec:overview}

In this section, we introduce the system architecture of the proposed control architecture, shown in Fig. \ref{fig:controlArchi}. 

We leverage an online high-frequency push-recovery NMPC to predict the instability in a short prediction window in response to unknown external pushes. The NMPC takes current state as input,  including CoM position ${\bm p}_c \in \mathbb R^{3}$, CoM velocity $\dot{\bm p}_c \in \mathbb R^{3}$, Euler angles $\bm \Theta \in \mathbb R^{3}$, angular velocity $\bm \omega \in \mathbb R^{3}$, joint positions $\mathbf q_j \in \mathbb R^{18}$, and joint velocities $\dot{\mathbf q}_j \in \mathbb R^{18}$. These states are generalized as  $\mathbf q$ and $\dot{\mathbf q}$. In addition, the first-time-step WBD in NMPC leverages the feedback of ground reaction force and moments (GRFM) solutions from the locomotion MPC to generate accurate float-base acceleration to aid the prediction. Then the gait frequency adaptation planner analyzes the unstable timesteps from the prediction, gathers the body attitude change, and determines the gait frequency and contact schedule for improved recovery. 
Subsequently, the locomotion MPC executes the stepping strategy for motion control.  On the low level, the swing foot control leverages heuristic foot placement policy \cite{raibert1986legged} while the force-to-torque mapping is used to convert GRFM and swing forces to leg joint torque. The arm joints are commanded to stay in-place with joint-space PD controllers.

\section{Proposed Approach}
\label{sec:approach}

In this section, we introduce the proposed approaches in this work, including the dynamics models of the humanoid robot, the push-recovery NMPC, the posture-aware gait frequency adaptation planner, and the locomotion MPC.

\subsection{HECTOR Humanoid and Dynamics Modelling }
\label{subsec:hector}

In this work, we use the HECTOR V2 humanoid robot as our robot model and hardware platform. HECTOR V2 is a successor of the HECTOR humanoid platform introduced in \cite{li2023hector}. HECTOR V2 humanoid consists of 5-DoF legs with ankle pitch actuation and 4-DoF arms. Standing at 85 $\unit{cm}$ and weighing 16 $\unit{kg}$, the robot can output 67.0 $\unit{Nm}$ maximum torque at the knee joints.

HECTOR's full joint-space dynamics equation of motion is described as follows. The joint-space generalized states ${\mathbf{q}} \in \mathbb{R}^{24}$ include float-based states  $\mathbf q_{fb}$, and joint states $\mathbf q_j$,
\begin{align}
\label{eq:fullDynamics}
    \mathbf{H}(\mathbf{q})\ddot{\mathbf{q}} + \mathbf{C}(\mathbf{q}, \dot{\mathbf{q}}) = \mathbf{\Gamma} + \bm{J}_i(\mathbf{q})^\intercal \bm{\lambda}_i 
\end{align}

where $\mathbf{H} \in \mathbb{R}^{24 \times 24}$ is the mass-inertia matrix and $\mathbf{C} \in \mathbb{R}^{24}$ is the joint-space bias force term. $\mathbf{\Gamma} = [\mathbf{0}_6; \bm \tau_j]$ represents the actuation in the generalized coordinate. $\bm \lambda_i $ and $\bm{J}_i$ represent the external force applied to the system and its corresponding Jacobian matrix.

While equation (\ref{eq:fullDynamics}) is capable of capturing the whole-body dynamics of the robot including its interaction with the ground or external objects, the computation burden associated with such a model poses challenges to deploying it in a highly efficient online MPC scheme. The author's prior work \cite{li2021force} investigated the effectiveness of a reduced-order model in linear MPC for humanoid locomotion, by simplifying the full humanoid robot model as a single rigid-body (SRB) with constant mass $m$ and body moment of inertia (MoI). The model is actuated by GRFM at the $n$th contact point. The GRFM consists of both 3-D force $\mathbf{F}_n$ and 3-D moment $\mathbf{M}_n$ entries in the world frame.

To address the limitation of the strong assumption of constant MoI in SRB dynamics and enhance the posture-regulating feature in MPC, in this paper, we introduce a CSRB dynamics model. It assumes the centroidal momentum stays constant for the relatively short duration of the prediction. This assumption allows a relaxed computation burden compared to full centroidal dynamics, which is nonlinear and requires full joint states as optimization variables. The CSRB dynamics is described as
\begin{align}
    \label{eq:cd1}
    \dot{\bm {\hat h}} =\bm A_{\mathrm{fb}}(\mathbf q_{0})\ddot{\mathbf q}_{\mathrm{fb}} +\dot{\bm A}( \mathbf q_{0},\dot{\mathbf q}_{0} )\dot{\mathbf q}_{0} =\begin{bmatrix}
    \Sigma^1_{i = 0} \{ \bm r_{i}^{f} \times \mathbf F_i + \mathbf M_i  \}\\
    \Sigma^1_{i = 0} \mathbf F_i + m \bm g
\end{bmatrix}
\end{align}
where $\dot{\bm {\hat h}}$ is the approximated rate of change of the spatial momentum matrix. We select the float-based part of the Centroidal Momentum Matrix \cite{orin2013centroidal, wensing2016improved}, $\bm A \in \mathbb{R}^{6\times 24}$, in the approximation to allow explicit expression of acceleration of float-base states,
\begin{align}
\label{eq:cd2}
    \ddot{\mathbf q}_{\mathrm{fb}} =
    \begin{bmatrix}
        \dot {\bm \omega} \\ \ddot {\bm p}_c
    \end{bmatrix}
    =\bm A_{\mathrm{fb}}^{-1} \Biggl( \begin{bmatrix}
    \Sigma^1_{i = 0} \{ \bm r_{i}^{f} \times \mathbf F_i + \mathbf M_i  \}\\
    \Sigma^1_{i = 0} \mathbf F_i + m \bm g
\end{bmatrix} - \dot{\bm A}\dot{\mathbf q}_{0}\Biggl),
\end{align}
where $\bm r_{i}^{f}$ is the distance vector from CoM $\bm p_c$ to the center of foot $i$. The $0$ subscript denotes the use of first-time-step joint states in $\bm A_{\mathrm{fb}}\in \mathbb{R}^{6\times 6}$ and $\dot{\bm A}$ throughout the prediction of the short-term dynamics, neglecting the prediction of joint states and allowing linearization in the locomotion MPC. 


\begin{figure}[!t]
\vspace{0.2cm}
		\center
		\includegraphics[clip, trim=0cm 3cm 3cm 0cm, width=1\columnwidth]{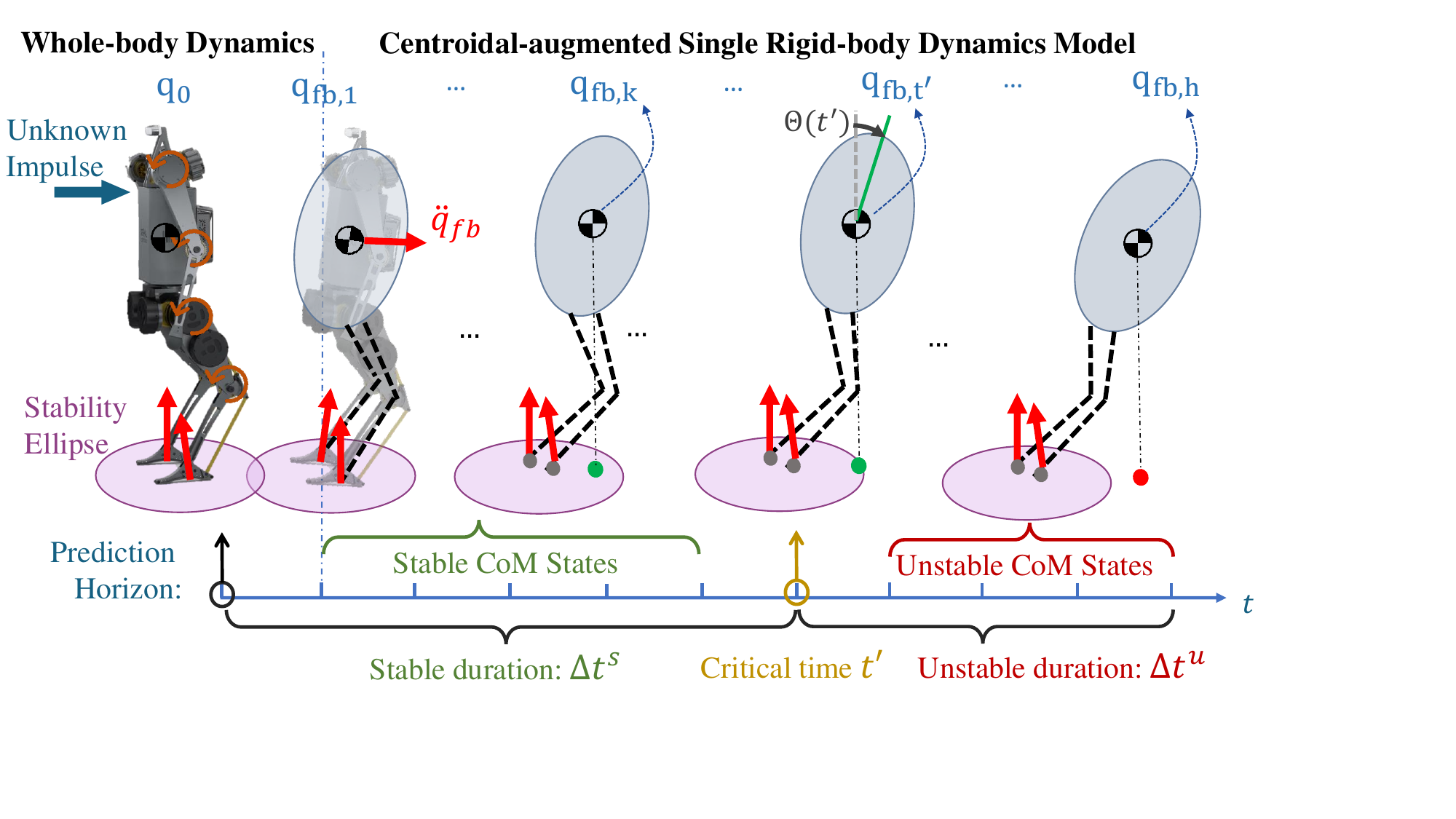}
		\caption{{\bfseries Illustration of CoM States Prediction under External Push}. With hierarchical-dynamics-based modelling in MPC}
		\label{fig:dynamics}
		\vspace{-0.4cm}
\end{figure}

\subsection{Push-recovery NMPC}
\label{subsec:NMPC}
This section presents the details of the proposed Push-recovery NMPC and its nonlinear programming problem formulation.

The Push-recovery NMPC is introduced to predict CoM state trajectory online and provide precursors of instability and large posture deviation for the subsequent gait frequency planner. Note that within the NMPC, the foot is assumed to remain on the ground during the prediction, allowing for a better assessment of the CoM trajectory.
To guarantee efficient online deployment, we leverage a hierarchical modeling strategy for MPC \cite{li2021model}. The WBD model of the robot is used for the first horizon of the NMPC and the CSRB dynamics with full kinematics is employed for the rest. We use the WBD model to accurately capture the robot float-base state evolution as a result of the external impulse applied to the robot at the first NMPC step, allowing a more accurate prediction of state evolution during the NMPC window.

The Push-recovery NMPC is formulated as a direct trajectory optimization problem. 
The optimization variables chosen in this problem are,
 \begin{align}
\label{eq:optvarNMPC}
\mathbf X = \{ \{\mathbf q_k, \:\dot{\mathbf q}_k\}^{h}_{k = 1}, \{\bm u_k\}^{h-1}_{k = 1}, \: dt^{\mathrm{nmpc}} \}; 
\end{align}
where $\bm u = [\mathbf F_0; \:\mathbf F_1; \:\mathbf M_0; \:\mathbf M_1]$ is the GRFM acting on the robot and $dt^{\mathrm{nmpc}}$ is the NMPC step length. The prediction horizon $h$ is synchronized with the prediction length of the locomotion MPC. The optimal control problem of the push-recovery NMPC is formulated as follows,
\begin{alignat}{3}
\label{eq:cost}
\min_{\mathbf X} \quad & \sum_{k = 0}^{h-1}\Bigl( \small\| \mathbf q_k -\mathbf q_0 \small\|^2 _{\bm Q_k} 
+  \small\| \bm u_k  \small\|^2 _{\bm R_k}\Bigr) \\ &+ \small\|\mathbf q_h -\mathbf q_0 \small\|^2_{\bm Q_h}+ \alpha (dt^{\mathrm{nmpc}}-dt_0)^2 \\ 
    \nonumber
    \textrm{subject to}  \quad & \quad
\end{alignat}
\vspace{-0.25cm}
\begin{subequations}
\allowdisplaybreaks
\setlength\abovedisplayskip{-3pt}
\begin{alignat}{3}
    \label{eq:qrange}
    \textrm{Joint Limits:} \quad & \mathbf q_{\mathrm{min}} \leq \mathbf q_{j,k} \leq \mathbf q_{\mathrm{max}}  \\
    \label{eq:FK}
    \textrm{Forward Kinematics:} \quad & \bm p^{f} = \texttt{FK}(\mathbf q_{k})\\
    \label{eq:sampletime}
    \textrm{Sampling Time:} \quad & 0.03\unit{s} \leq {dt}^{\mathrm{nmpc}} \leq 0.05\unit{s} \\
    \label{eq:GRF}
    \textrm{Force Limit:} \quad & 0 \leq F_{z} \leq F_{\mathrm{max}}\\
    \label{eq:friction}
    \textrm{Friction Cone:} \quad & \sqrt{F_{x}^2 + F_{y}^2} \leq \mu F_{z} \\
    \label{eq:linefoot}
    \textrm{Line foot:} \quad & -l_\textrm{h}  F_{z} \leq M_{y} \leq l_\textrm{t}  F_{z} 
\end{alignat}
\end{subequations}
\begin{alignat}{3}
    \nonumber
     \textrm{WBD (\ref{eq:fullDynamics}) when} \: k = 0: \quad \quad \quad \quad & \\
    \label{eq:wbd}
    \{\dot{\mathbf q}_1,\: \mathbf q_1\} = \bm f_{\mathrm{WBD}}(\mathbf q_0,\:\dot{\mathbf q}_0,\:& \bm u_{0},\: {dt}^{\mathrm{nmpc}})
\end{alignat}
\begin{alignat}{3}
    \nonumber
    \quad  \textrm{CSRB Dynamics}&\textrm{ (\ref{eq:cd2}) when}   \: 0< k \leq h-1:\\
    \label{eq:flightDynamicsCons}
    \{\dot{\mathbf{q}}_{\mathrm{fb},k+1},\:\mathbf{q}_{\mathrm{fb},k+1}\}& = \bm f_{\mathrm{CSRB}}(\mathbf{q}_{\mathrm{fb},k},\:\dot{\mathbf{q}}_{\mathrm{fb},k},\:\bm{u}_{k},\: {dt}^{\mathrm{nmpc}}) 
\end{alignat}

The objectives include regulating the CoM states to stay close to the initial condition $\mathbf q_0$, minimizing GRFM, and penalizing the sample time to a nominal value $dt_0 = 0.04$s, weighted by $\bm Q\in \mathbb R^{24\times 24}$, $\bm R\in \mathbb R^{12\times 12}$, and $\alpha\in \mathbb R$. 
 
Equations (\ref{eq:qrange}-\ref{eq:FK}) are the hardware kinematics constraints. The current foot position $\bm p^f$ in the world frame is assumed stationary during stance. Forward kinematics (\texttt{FK}) is performed to obtain the current joint angles. 
Equation (\ref{eq:GRF}-\ref{eq:linefoot}) represents the GRFM constraints, including contact friction cone, normal force limit ($F_z = 250 \unit{N}$) and line-foot constraints, where $\mu$ is the ground friction coefficient, $l_\textrm{h}$ and $l_\textrm{t}$ are the line-foot parameters.

Additionally, we incorporate the sample time,  $dt^\textrm{nmpc}$, as an optimization variable to relax kinematic constraints. Under a strong push, when far-horizon kinematics are violated with a nominal sample time, due to the float-base SRB dynamic evolution disregarding these constraints, optimizing $dt^\textrm{nmpc}$ provides flexibility to shorten the prediction window and maintain feasibility. This approach also facilitates variable-frequency walking, as the gait frequency adaptation law directly depends on the optimization decision of $dt^\textrm{nmpc}$.

\subsection{Posture-aware Gait Frequency Adaptation Planner}
\label{subsec:SSP}

In this section, we introduce a gait frequency adaptation law to plan for humanoid stepping frequency for improved push-recovery performance and posture regulation. It analyzes the duration of unstable states and body attitude change in the NMPC prediction. Note that we attempt to primarily leverage gait frequency to achieve these improved behaviors instead of planning for foot placement concurrently. 

First, we approximate the stability region of the humanoid as an ellipse. Consistent throughout the prediction horizon, the region is described as an elliptical area with its center placed in the center of the polygon formed by the foot contact locations at $k=0$. The ellipse axis lengths are determined by the maximum recoverable CoM displacement with double-support stance. At NMPC prediction step $k$, if the projection of the CoM on the ground falls within this stability region, it is classified as a stable time step; otherwise, it is considered an unstable one. The entire prediction horizon is then divided into stable and unstable segments, as illustrated in Figure \ref{fig:dynamics}. 
The duration of each segment is,
 \begin{align}
\label{eq:stabRegi}
\Delta t^s = dt^{\mathrm{nmpc}}\cdot h^s,\: 0 < h^s \leq h,\\
\Delta t^u = dt^{\mathrm{nmpc}}\cdot h^u,\: 0 \leq h^u < h,
\end{align}
where $h^s$ and $h^u$ are the number of stable and unstable horizons. 
Next, we introduce gait frequency adaptation laws to determine the swing duration $\Delta t'$ in the locomotion MPC, based on NMPC predictions. We proposed two laws, 
 \begin{align}
\label{eq:swingDuration}
\textrm{Law 1:}\: \Delta t' = \beta \cdot \Delta t^u + \gamma \cdot e(\bm q(0), \bm q(t')),\\
\textrm{Law 2:}\: \Delta t' = \beta \cdot \Delta t^s - \gamma \cdot e(\bm q(0), \bm q(t')), \\
e(\bm q(0), \bm q(t')) = 1 - | \bm q(0)\bm q(t') |,
\end{align}
where $\beta$ and $\gamma$ are scaling parameters. 
Gait frequency law 1 permits a longer swing time under large impulses, allowing for correction of the landing position and enabling a larger stride to maintain balance. On the other hand, law 2 favors faster stepping in response to sudden large body accelerations, similar to the learning-based behaviors observed in \cite{van2024revisiting}. We also add consideration of body attitude deviation at the critical time $t'$ in the NMPC prediction, where $e$ represents the body quaternion $\bm q$'s error map between attitudes of an upright body (\textit{i.e.}, $\bm q(0) = [1\ 0\ 0\ 0]$) and predicted attitude at the critical time. (\textit{e.g.}, y-direction moment impulse may cause pitch angle deviation).
We conduct comparative analyses of both laws in Section \ref{sec:Results}.

The stepping strategy and planner activate upon detecting unstable CoM states or significant body orientation changes, shifting the locomotion MPC to stepping mode. While stepping, the NMPC and planner are inactivated until the completion of the entire two-step gait cycle, they will then determine if stability is reached, or else stepping will resume based on a newly computed gait frequency. The planner overwrites the contact sequence by selecting the first swing leg upon switching to stepping based on the final CoM's $y$-position sign to better counteract instability.


\begin{figure*}[!t]
\vspace{0.2cm}
		\center
		\includegraphics[clip, trim=0cm 6.25cm 0.5cm 0cm, width=2\columnwidth]{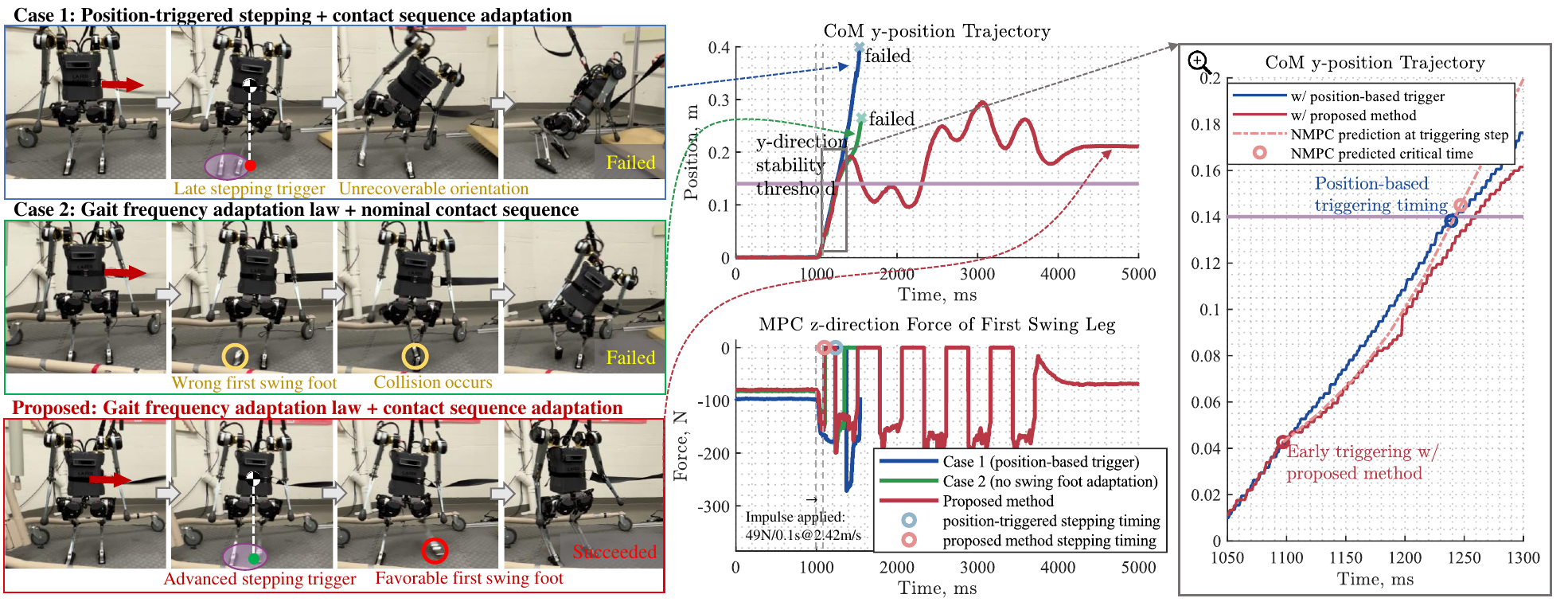}
		\caption{{\bfseries Push-recovery Comparative Experiment Snapshots with Baseline Methods}. Snapshots and plots of NMPC-prediction-based stepping trigger vs. current-position-based stepping trigger. CoM $y$-position plot is zoomed in to showcase the prediction trajectory of NMPC for time-advanced stepping triggering. }
		\label{fig:hardware2}
		\vspace{-0.2cm}
\end{figure*}


\subsection{Locomotion MPC for Push-recovery Balancing}

In the lower-hierarchy locomotion MPC, we integrate the proposed CSRB dynamics model in a convex MPC for real-time high-frequency motion control. Though not a main contribution of the paper, we aim to use this model to enhance the previous single-rigid-body-based MPC (SRBM-MPC) \cite{li2021force}.

Upon the detection of instability by the planners,  the computed swing duration $\Delta t'$ is then used to set the time step length $dt^{\mathrm{mpc}}$ for the MPC, and 
\begin{align}
dt^{\mathrm{mpc}} = \frac{\Delta t'}{h_{\mathrm{stance}}}, \  0.04\unit{s} \leq dt^{\mathrm{mpc}} \leq 0.065\unit{s},
\end{align}
where $h_{\mathrm{stance}}$ represents the number of discrete time steps in the MPC that constitute an entire swing phase. In this study, we use $h_{\mathrm{stance}} = 5$, and the range of $dt^{\mathrm{mpc}}$ is based on hardware-achievable gait frequencies.

The MPC formulation is similar to the author's prior work \cite{li2023hector}, with a newly introduced CSRB dynamics to replace the original SRB. In the state-space model, we choose to include the robot CoM states and scalar 1 in the optimization variable $\bm x = [\mathbf \Theta;\: \bm p_c;\: \bm \omega;\: \dot{\bm p}_c; \: 1] \in \mathbb{R}^{13}$ to linearize the dynamics and form $\dot{{\bm { x}}}(t) =  {\hat{\mathbf A}_c} {{\bm {x}(t)}} +  {\hat{\mathbf B}_c} \bm u(t)$, where ${\hat{\mathbf A}_c \in \mathbb{R}^{15\times13}}$ and ${\hat{\mathbf B}_c \in \mathbb{R}^{13\times12}}$ are continuous-time state-space dynamics matrices. At time step $k$, we discretize the state-space dynamics with step length $dt^{mpc}$: 
\begin{align}
\label{eq:discreteSS}
\bm {x}_{k+1} = \mathbf {\hat{A}}_k\bm x_k + \mathbf {\hat{B}}_k\bm u_k
\end{align}
\begin{align}
\label{eq:discreteAB}
\hat{\mathbf A}_k = \mathbf I_{13} + \hat{\mathbf A}_c(t)\cdot dt^{\mathrm{mpc}},\quad \hat{\mathbf B}_k =  \hat{\mathbf B}_c(t)\cdot dt^{\mathrm{mpc}}
\end{align}
\begin{align}
    \hat{\mathbf A}_c = 
    \begin{bmatrix}
        \mathbf 0_3 & \mathbf 0_3 & \mathbf S_R^{-1} & \mathbf 0_3 & \mathbf 0_{3\times1}\\
        \mathbf 0_3 & \mathbf 0_3 & \mathbf 0_3 & \mathbf I_3 & \mathbf 0_{3\times1}\\
        \mathbf 0_{6\times3} &  \dots &  \dots & \mathbf 0_{6\times3} & \bm A_{\mathrm{fb}}^{-1}(\begin{bmatrix}
            \bm 0 \\ m \bm g
        \end{bmatrix}-\dot{\bm A}\dot{\mathbf q}_{0}) \\
        \mathbf 0_{1\times3} &  \dots &  \dots & \mathbf 0_{1\times3} & 0
    \end{bmatrix},
\end{align}
\begin{align}
    \hat{\mathbf B}_c = 
    \begin{bmatrix}
      \mathbf{I}_6 &  \dots & 0\\
       \vdots &  \bm A_{\mathrm{fb}}^{-1} & \vdots\\
       0 & \dots & 1
    \end{bmatrix}
    \begin{bmatrix}
        \mathbf 0_3 &  \dots &  \dots & \mathbf 0_3 \\
        \mathbf 0_3 &  \dots &  \dots & \mathbf 0_3 \\
        [\bm r_0^f]_\times & [\bm r_1^f]_\times & \mathbf I_3 & \mathbf I_3 \\
        \mathbf I_3 & \mathbf I_3 & \mathbf 0_3 & \mathbf 0_3 \\
        0 & \dots & \dots & 0
    \end{bmatrix},
\end{align}
where $\mathbf S_R$ is the mapping between the rate of change of Euler angles and body angular velocity, followed from \cite{li2023hector}. 

The convex MPC finite-horizon optimization problem, swing-foot heuristic policy, and low-level force-to-torque mappings follow standard techniques which are explained in detail in prior work \cite{li2023hector}.

\section{Results}
\label{sec:Results}

This section presents comparative analyses based on numerical results and hardware experimentation.

\subsection{Validation Setup}
For numerical validations, we use a MATLAB/Simulink simulation based on the HECTOR open-source project \cite{hectorGithub}. The NMPC planner is solved via \texttt{fatrop} solver \cite{vanroye2023fatrop} with CasADi toolbox \cite{Andersson2019}. The locomotion MPC is solved via \texttt{qpOASES} solver \cite{ferreau2014qpoases}. In hardware experiments, the NMPC is code-generated through CasADi Python interface to C++ environment and is deployed in a custom ROS-based interface for real-time hardware control. We use the \texttt{Pinnochio} library \cite{carpentier2019pinocchio} to compute whole-body forward dynamics efficiently. The onboard computer to execute the proposed control scheme in real-time is equipped with an Intel i5 1340P Processor. The push-recovery NMPC's solve time averages around 10 \unit{ms} and the locomotion MPC's solve time averages under 1 \unit{ms}. 

\begin{figure}[!t]
\vspace{0cm}
     \centering
     \begin{subfigure}[b]{0.48\textwidth}
        \centering
        \includegraphics[clip, trim=1.5cm 9.5cm 1cm 9.5cm, width=1\columnwidth]{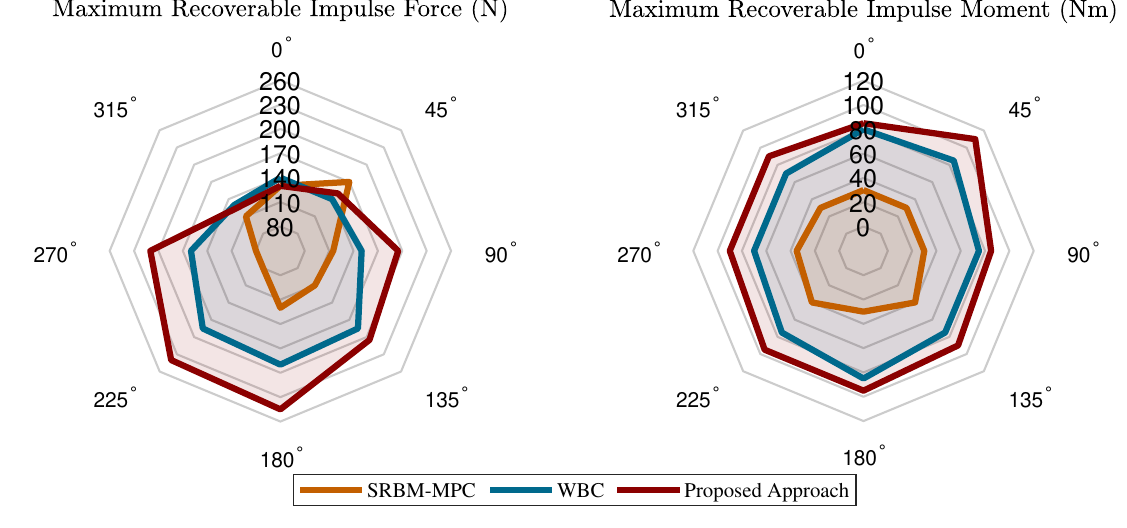}
     \caption{Comparing with baseline control approaches. }
     \label{fig:forceMoment_radar_map}
     \end{subfigure}
     \begin{subfigure}[b]{0.48\textwidth}
         \centering
	   \includegraphics[clip, trim=1.5cm 9.5cm 1cm 9.0cm, width=1\columnwidth]{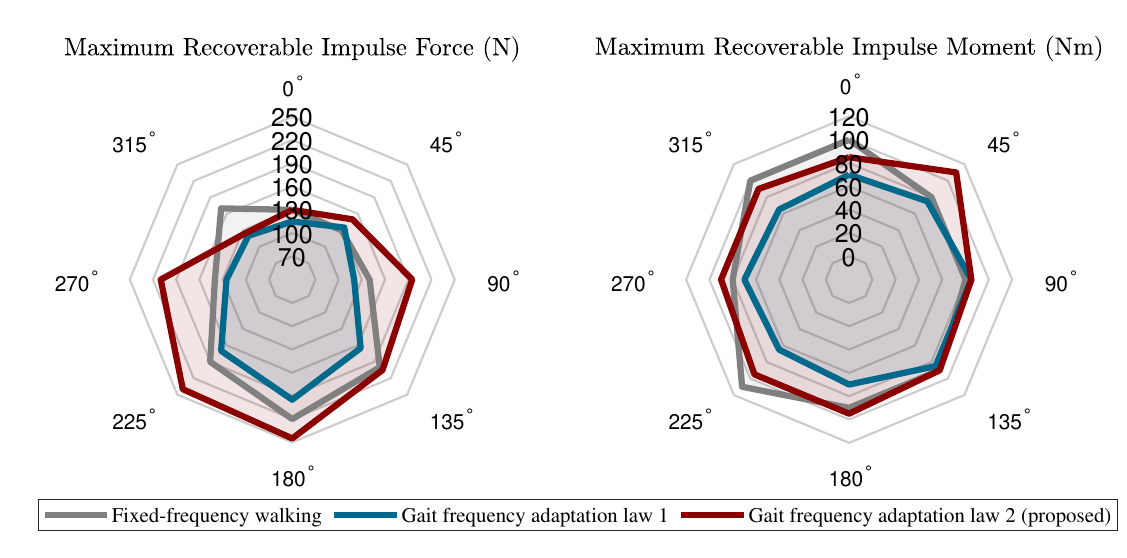}
     \caption{Comparing stepping-frequency laws. }
        \label{fig:dt_radar_map}
     \end{subfigure}
     \caption{{\bfseries{Maximum Recoverable External Force and Moment Impulse Radar Plots in 8 Directions.}} Comparative analysis of (a) with different control approaches and (b) with different stepping frequency laws.}
        \label{fig:comparisons}
        \vspace{-0.2cm}
\end{figure}

\begin{figure*}[!t]
\vspace{0.2cm}
		\center
		\includegraphics[clip, trim=0cm 6.1cm 0cm 0cm, width=2\columnwidth]{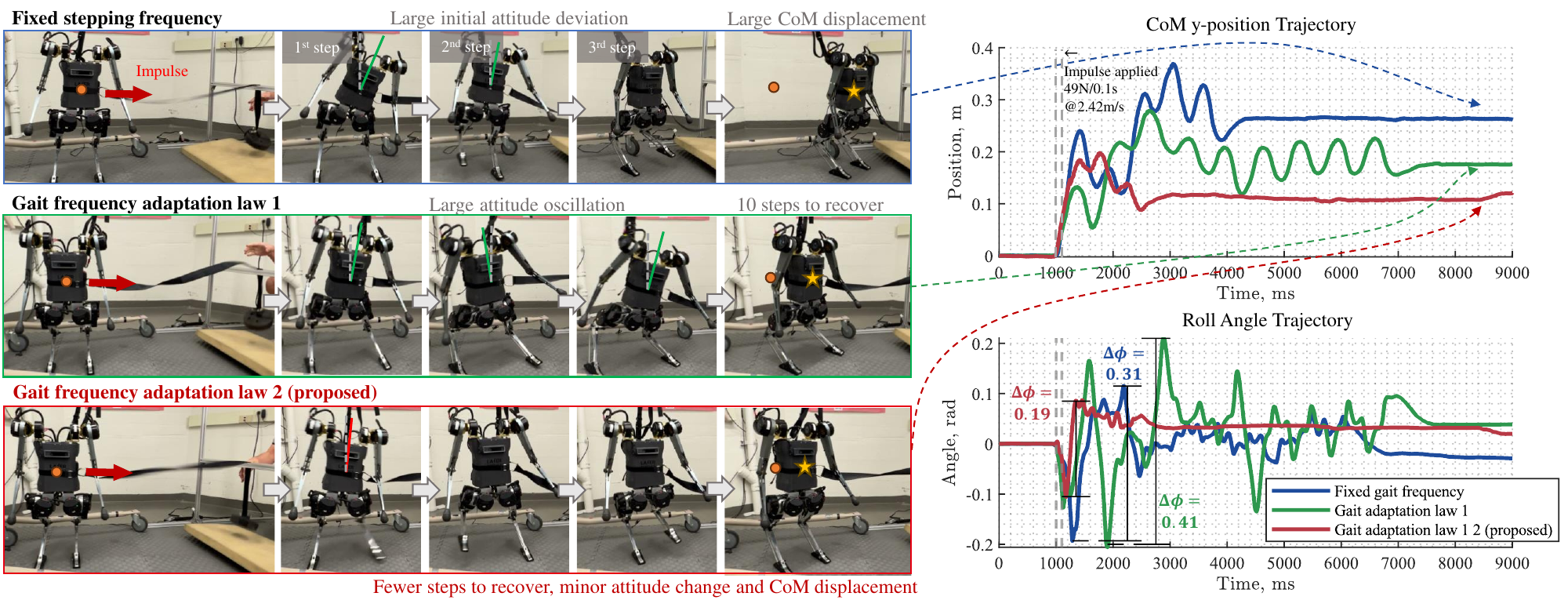}
		\caption{{\bfseries Push-recovery Comparative Experiment Snapshots with Gait Frequency Adaptation Laws}. Snapshots and plots illustrate the comparisons of $y$-direction CoM and body roll angle trajectory using (1) fixed gait frequency, (2) law 1: favoring slow gait frequency, and (3) law 2: favoring fast gait frequency. The orange dot and yellow star in each experiment represent the robot's initial and final CoM locations.  }
		\label{fig:hardware1_plots}
		\vspace{-0.4cm}
\end{figure*}

To maintain relatively consistent impulse magnitude and duration in the hardware experiments, we adopted a setup by \cite{van2024revisiting}, as illustrated in Fig. \ref{fig:controlArchi}. The robot's waist is connected to an inelastic rope with a free-falling weighted object at the other end to simulate a short-duration external impulse. Upon release, the object free-falls a distance of $h_1$ before the rope becomes taut, at which point an external force equivalent to the object's weight is applied to the robot, with a falling speed of $v_1$. The impulse continues until the weight reaches the ground after an additional distance of $h_2$. For consistency, we set $h_1 = 0.3\ \unit{m}$, $h_2 = 0.1\ \unit{m}$, and the object's mass $m_{\mathrm{ext}} = 5.0\ \unit{kg}$. By calculation, the object reaches an impact velocity of $v_1 = 2.42\ \unit{m/s}$. The impulse duration, recorded using a high-speed camera, is approximately $0.1\ \unit{s}$.

\subsection{Comparative Analyses in Numerical Simulation}
To validate the effectiveness of the proposed control scheme in push recovery, we first compare the maximum allowable external force and moment impulses on the robot with the baseline approaches in simulation, including 
\begin{enumerate}
    \item SRBM-MPC \cite{li2023hector},
    \item Whole-body Control (WBC) \cite{wensing2013generation,ding2022orientation},
    \item Proposed hierarchical-MPC-based approach.
\end{enumerate}
The impulses are applied for 0.2 s in stance mode, tested in 8 directions, where $0^\circ$ represents the force or moment applied in the $+x$ direction of the robot trunk. For baseline approaches (1) and (2), the \textit{stepping} strategy initiates once the current robot CoM position is outside of the support region (\textit{i.e.}, position-based trigger). 

As illustrated in Fig. \ref{fig:forceMoment_radar_map}, with our proposed approach, the magnitudes of the maximum recoverable impulses in terms of both external force and moment surpass the baseline approaches. The average improvement in all directions is 131$\%$ compared with SRBM-MPC and 18.1$\%$ compared with WBC running at 1 kHz.

Secondly, the gait frequency of the recovery steps significantly impacts the robot's stability and posture during recovery. In this comparative analysis, 
as described in Section \ref{subsec:SSP}, we compare push-recovery performance using:
\begin{enumerate}
    \item Fixed gait frequency,
    \item Law 1: gait frequency favoring slow frequency,
    \item Law 2: gait frequency favoring fast frequency.
\end{enumerate}
All methods are integrated with the proposed push-recovery NMPC and locomotion MPC, but employ different strategies in the stepping planner to determine stepping frequencies for the recovery steps. Fig. \ref{fig:dt_radar_map} shows the maximum recoverable external force and moment impulses using these approaches. It is noticed that our proposed gait frequency adaptation law 2 allows larger allowable impulses compared to others, particularly with force impulses.

\subsection{Comparative Analyses through Hardware Validation}
Next, we validate the effectiveness of our proposed scheme in hardware experiments, to demonstrate (1) the importance of push-recovery NMPC + gait frequency adaptation in advancing stepping timing, (2) improved push-recovery performance, and (3) minimized body attitude change.

Firstly, we demonstrate the significance of our proposed framework in advancing the stepping timing for improved push-recovery performance. We present experimental comparisons of (1) using position-triggered stepping law, (2) push-recovery stepping without contact-sequence adaptation, and (3) the proposed method. 
Fig.~\ref{fig:hardware2} shows the snapshots of these experiments.
Using a position-based stepping trigger often delays the robot's recovery. Plots comparing CoM $y$-position for position-based vs. NMPC-prediction-based triggers show that with our proposed scheme, stepping is initiated 125 ms earlier, as seen in the MPC $z$-force plots where zero force indicates swing phase.

We then conduct a comparative analysis of the performance of posture regulation in push-recovery with the three compared stepping strategies in Fig. \ref{fig:dt_radar_map}.
The experimental snapshots are shown in Fig.~\ref{fig:hardware1_plots}. It is observed that the slow gait frequency (law 1) produces large roll angle difference ($\Delta \phi = 0.41 $ rad) and oscillation during each step, resulting in an excessive number of steps to recover balance. In the case of nominal fixed-frequency walking, the CoM $y$ position has a large displacement of 0.26 m. With the proposed gait frequency adaptation law 2, we observe a superior performance in both body attitude regulation ($\Delta \phi = 0.19 $ rad)  and CoM displacement regulation ($\Delta {p}_{c,y} = 0.11$ m).

\section{Conclusions}
\label{sec:Conclusion}

In conclusion, we propose a hierarchical-MPC-based framework for improved performance and body-attitude regulation through varying gait frequencies during recovery stepping on humanoid robots under unknown disturbances. Our framework consists of a hierarchical-model-based nonlinear MPC for CoM trajectory prediction, a posture-aware gait frequency adaptation law to determine stepping strategies, and a locomotion MPC to execute rebalancing stepping with tailored gait frequency. In both simulation and hardware validations, under unknown impulses, our proposed NMPC and gait frequency adaptation law showcases the advantage of handling larger unknown external impulses and achieving less body attitude deviation.

\newpage
\balance
\bibliographystyle{ieeetr}
\bibliography{reference.bib}

\end{document}